\documentclass[journal,comsoc]{IEEEtran}
\usepackage{graphicx}
\usepackage{amsmath}
\usepackage{pifont}
\newcommand{\cmark}{\ding{51}}%
\newcommand{\xmark}{\ding{55}}%
\usepackage{multirow}
\usepackage{booktabs}
\usepackage{float}
\usepackage{graphics} 
\usepackage{mathptmx} 
\usepackage{times} 
\usepackage{threeparttable}

\usepackage{hyperref}
\hypersetup{
    colorlinks=false,
    linkcolor=magenta,
    filecolor=magenta,      
    urlcolor=magenta,
}

\usepackage[T1]{fontenc}

\ifCLASSINFOpdf
\else
\fi
\usepackage{amsmath}
\begin{document}
%

\title{ACM-Net: Action Context Modeling Network for Weakly-Supervised Temporal Action Localization}

\author{Sanqing~Qu$^{1}$,~Guang~Chen$^{1, 2*}$,~\IEEEmembership{Member,~IEEE,} ~Zhijun~Li$^{3}$,~\IEEEmembership{Senior Member,~IEEE,} ~Lijun~Zhang$^{1}$, ~Fan~Lu$^{1}$,  ~Alois~Knoll,$^{2}$~\IEEEmembership{Senior Member,~IEEE}
\thanks{$^*$Corresponding author:  Guang~Chen, Email: guangchen@tongji.edu.cn}
\thanks{Authors Affiliation: $^1$Department of Automotive Engineering, Tongji University, Shanghai, China; $^2$Robotics, Artificial Intelligence and Embedded Systems, Technical University of Munich, Munich, Germany;
$^3$Department of Automation, University of Science and Technology of China, China;}}

\markboth{IEEE TRANSACTIONS ON Image Processing}%
{Shell \MakeLowercase{\textit{et al.}}: Bare Demo of IEEEtran.cls for IEEE Communications Society Journals}
%



\maketitle

\begin{abstract}
\par Weakly-supervised temporal action localization aims to localize action instances temporal boundary and identify the corresponding action category with only video-level labels. Traditional methods mainly focus on foreground and background frames separation with only a single attention branch and class activation sequence. However, we argue that apart from the distinctive foreground and background frames there are plenty of semantically ambiguous action context frames. It does not make sense to group those context frames to the same background class since they are semantically related to a specific action category. Consequently, it is challenging to suppress action context frames with only a single class activation sequence. To address this issue, in this paper, we propose an action-context modeling network termed ACM-Net, which integrates a three-branch attention module to measure the likelihood of each temporal point being action instance, context, or non-action background, simultaneously. Then based on the obtained three-branch attention values, we construct three-branch class activation sequences to represent the action instances, contexts, and non-action backgrounds, individually. To evaluate the effectiveness of our ACM-Net, we conduct extensive experiments on two benchmark datasets, THUMOS-14 and ActivityNet-1.3. The experiments show that our method can outperform current state-of-the-art methods, and even achieve comparable performance with fully-supervised methods. Code can be found at \url{https://github.com/ispc-lab/ACM-Net}.
\end{abstract}

\begin{IEEEkeywords}
weakly-supervised learning, temporal action localization, action-context modeling network
\end{IEEEkeywords}

%
 
\section{Introduction}

\par \IEEEPARstart{W}{ith} the explosive growth of video contents, understanding and learning from videos has attracted great interest in computer vision community. As one of the fundamental but challenging tasks of video understanding, temporal action localization or detection that aims to localize and classifying action instances in untrimmed videos has drawn lots of attention, due to its great potential for video retrieval~\cite{video_retrieval_1, video_retrieval_2}, summarization~\cite{video_summarization_1}, surveillance~\cite{video_surveillance_1, video_surveillance_2}, anomaly detection~\cite{video_anomaly_detection} and more. Thanks to the rapid development of deep learning, recently, a plenty of methods~\cite{SSN, BSN, BMN, BSN++} have been proposed and achieved remarkable performance under the fully supervised definition. However, these methods require precise temporal annotation of each action instance during training, which is time-consuming, error-prone and extremely costly to collect. In contrast, weakly supervised temporal action localization (W-TAL), which requires only video-level action category labels, is a more reasonable choice and has attracted a great deal of attention. Compared with precise temporal boundary annotations of action instances, the video-level action category label is easier to collect and beneficial to avoid localization bias introduced by human annotators.

\par Existing weakly supervised temporal action localization approaches can be divided into two main categories. One kind of approaches~\cite{STPN, A2CLPT, TSM, background_modeling, IWONet}, inspired by the weakly-supervised image semantic segmentation task~\cite{weakly_seg_1, weakly_seg_2, weakly_seg_3}, formulate the weakly-supervised temporal action localization as a video recognition problem and introduce a foreground-background separation attention mechanism to construct video-level features, then apply an action classifier to recognize videos. While the other approaches~\cite{WTALC, MAAN, BASNet, EMMIL} formulate this problem as a multi-instance learning task~\cite{MIL} and treat the entire untrimmed video as a bag containing both positive and negative instances, i.e. foreground action instances frames and background non-action frames. These methods first apply a classifier to obtain temporal-point class activation sequence (CAS) and then employ a top-k mechanism to aggregate the video-level classification scores. 

\par
\begin{figure*}[ht]
    \centering
    \includegraphics[width=1.00\textwidth]{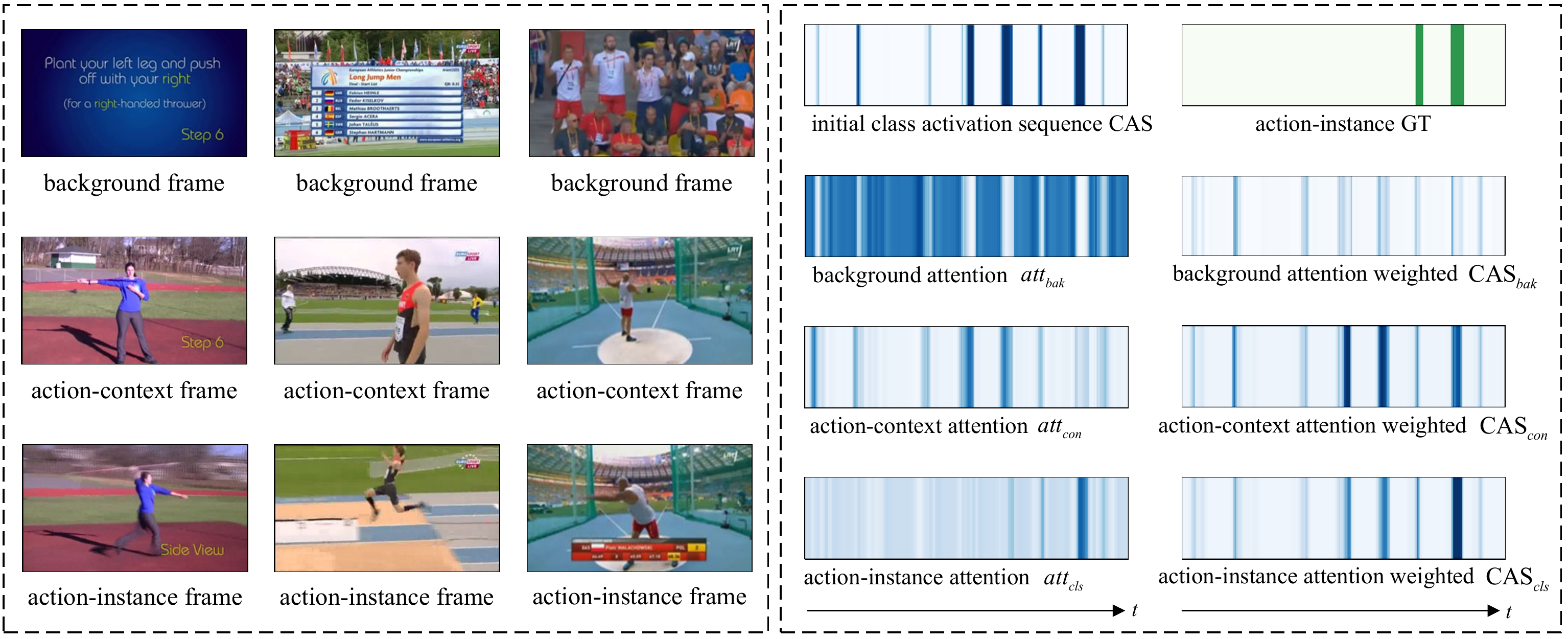}
    \caption{Apart from distinctive action instances and non-action background frames, there are plenty of semantically ambiguous action context frames. Traditional weakly-supervised temporal localization methods mainly apply foreground-background attention mechanism to separate action instances frames and non-action frames (non-action background and semantically ambiguous action context frames). However, these methods are not capable of suppressing those context frames well, since they are semantically related to specific action instances, which makes no sense to directly assign these frames to the background class. To address this issue, we propose a three-branch attention module to measure the likelihood of each temporal point being action instances, action contexts, or non-action background. Based on the obtained attention values, we then construct three-branch class activation sequence for action instances, contexts, and non-action background, respectively. As we can see in the above figure, this mechanism is greatly beneficial for us to suppress those semantically ambiguous context frames.}
    \label{fig:motivation}
\end{figure*}

 As can be observed from the above discussion, both types of methods aim at learning the effective classification functions to identify action instances from bags of action instances and non-action frames. However, there is still a huge performance gap between the weakly supervised and supervised methods. We argue that the reason may lie in the fact that the untrimmed video contains a number of semantically ambiguous action context frames in addition to the discriminative foreground action instances and static non-action background frames. As we illustrated in Fig.~\ref{fig:motivation}, it is challenging to distinguish action instances and action contexts based on a simply foreground-background separation CAS since it does not make sense to assign these context frames directly to the background class due to that they are action-related and do not share the same semantic information as the contexts of other action categories.

\par To realize action contexts suppression under video-level supervision, in this paper, we propose an action context modeling network (ACM-Net). Specifically, we first introduce a classification branch to obtain the initial class activation sequence (CAS). But as we mentioned before, this initial CAS is not capable of suppressing ambiguous action context frames, since they are action semantically related. To address this issue, we propose a three-branch class-agnostic attention module to discriminate action instances, action contexts, and non-action backgrounds, individually. Then based on these three-branch attention values, we construct new three-branch class activation sequence, i.e. $\mathrm{CAS}_{ins}$, $\mathrm{CAS}_{con}$, $\mathrm{CAS}_{bak}$, which denotes attention values weighted action instances CAS, action contexts CAS, and the background CAS, respectively. Thereafter, we apply the multi-instance learning mechanism to compute video-level classification scores to realize separation among action instances, action contexts, and non-action background. The detailed framework is presented in Fig.~\ref{fig:framework}. To validate the effectiveness of our ACM-Net, we conduct extensive experiments on the THUMOS-14~\cite{thu14_dataset} and the ActivityNet-1.3~\cite{Activitynet} datasets. The results show that our ACM-Net can successfully realize the separation of action instances and action contexts, and achieve new state-of-the-art performance on both benchmarks.

\par The main contribution are summarized as three-fold:
\begin{itemize}
    \item Different from the previous methods that divide  the frames of a video into the only foreground and background frames, we argue that there are some semantically ambiguous action context frames. In this paper, we investigate how action context modeling will affect the weakly-supervised temporal action localization and propose an action context modeling network (ACM-Net) to realize the separation of action contexts and action instances.

    \item The proposed ACM-Net integrates a class agnostic three-branch attention module to measure the likelihood of each temporal point containing action instance, action context, and non-action background frames simultaneously. Based on the obtained attention values, we then construct three-branch class activation sequences to achieve the distinguishing of action instances, action contexts, and non-action backgrounds.
    
    \item We conduct extensive experiments on the THUMOS-14 and ActivityNet-1.3 datasets. The qualitative visualization results demonstrate the effectiveness of our ACM-Net in distinguishing between ambiguous action instances and action instances. And the quantitative results show that our ACM-Net outperforms current state-of-the-art methods and even can achieve comparable performance with recently fully-supervised methods.
\end{itemize}

\section{Related Work}

\subsection{Action Recognition}
\par As one of the fundamental tasks of video understanding, action recognition aimed at identifying actions in trimmed video clips has been extensively studied. Early approaches~\cite{IDT, fisher_recognition} mainly focused on design effective hand-crafted descriptors that incorporate spatial-temporal features. In recent years, with the development of deep learning, there are plenty of networks have been proposed. These methods are mainly constructed based on image-level backbone networks~\cite{AlexNet, GoogleNet, ResNet}. Early approaches~\cite{two_stream, TSN, TRN} directly apply these image backbone networks to RGB and optical flow images to model the spatial temporal information. To further improve recognition performance, researchers extend 2D convolution operation to 3D by extending the temporal dimension, and 3D CNN based models~\cite{C3D, I3D, I2D, R3D, 3DCNN} (including 2D spatial convolution plus 1D temporal convolution) become the mainstream methods. However, although these methods achieve significant performance on trimmed video clips, in practice, long untrimmed videos are more often encountered, which makes these methods unable to achieve accurate semantic information modeling and limits the practical application.

\subsection{Fully-Supervised Temporal Action Localization}
\par Different from action recognition, which focuses only on trimmed video clips to identify action categories, the temporal action localization task aims not only to classify action instances but also to localize the start and end temporal boundary of action instances in long untrimmed videos. Temporal action localization with full supervision requires manual annotated temporal boundary and category of each action instance in videos during training. Inspired by 2D object detection, a plenty of methods~\cite{TCNN,CDC,TURN_TAP,SSN,BSN,rethinking_faster_rcnn,BMN,GTAD,BSN++} adopt a two-stage paradigm, i.e. proposal generation and classification. Given full action instances annotations, two-stage methods usually filter out the non-action proposals at the proposal generation stage by introducing a binary classifier and then introduce temporal feature modeling to realize action proposals classification and boundary refinement. Currently, there are two main categories of proposal generation methods, namely top-down framework~\cite{TCNN,CDC,TURN_TAP,rethinking_faster_rcnn,GTAD} and bottom-up framework~\cite{SSN,BSN,BMN,BSN++}. The former methods usually generate proposals with pre-defined regular distributed segments, e.g. sliding windows based methods, which is not flexible and often causes extensive false positive proposals. To address the above issues, the latter methods train a detector to search specific action points, e.g, action boundary or center points, and then combine these points to generate action proposals. However, since all methods require action instance labels during the proposal generation and classification stage, they are inevitably causing heavy annotation costs and not capable of widely employed in reality.

\subsection{Weakly-Supervised Temporal Action Localization}
\par The weakly-supervised temporal action localization methods are proposed to reduce the expensive annotation costs, compared with the supervised temporal action localization that needs precise annotation of each action instance, during training, the weakly-supervised temporal action localization methods require only video-level action category labels. Existing  weakly-supervised  temporal action localization approaches can be divided into two branches. Inspired by the weakly-supervised image semantic segmentation task~\cite{weakly_seg_1, weakly_seg_2, weakly_seg_3}, the first framework methods~\cite{STPN, A2CLPT, TSM, background_modeling, IWONet} formulate this task as an action recognition problem and introduce a foreground-background separation attention branch to construct video-level features, then apply an action classifier to recognize videos. While the latter framework approaches~\cite{WTALC, MAAN, BASNet, EMMIL} formulate this problem as a multi-instance learning task~\cite{MIL} and treat the entire untrimmed video as a bag containing both positive and negative instances. They first obtain frame-level action recognition scores, i.e. the class activation sequence CAS, and then introduce a top-k  mechanism to construct video-level classification scores. However, though these methods have achieved significant performance, there is still a performance gap between fully-supervised methods, we attribute this to that there are plenty of ambiguous action context frames apart from distinctive action instances and non-action background frames. It's challenging based on only a single foreground-background separation mechanism to suppress those frames as they are semantically related to specific actions. To address this issue, we introduce an action context modeling network, namely ACM-Net, which integrates a three-branch attention module to measure the likelihood of each temporal point containing action contexts. And then we build three-branch class activation sequences CAS based on the obtained attention values to realize action instances, contexts, and non-action background frames separation.

\section{Methodology}
\begin{figure*}[ht]
    \centering
    \includegraphics[width=1.00\textwidth]{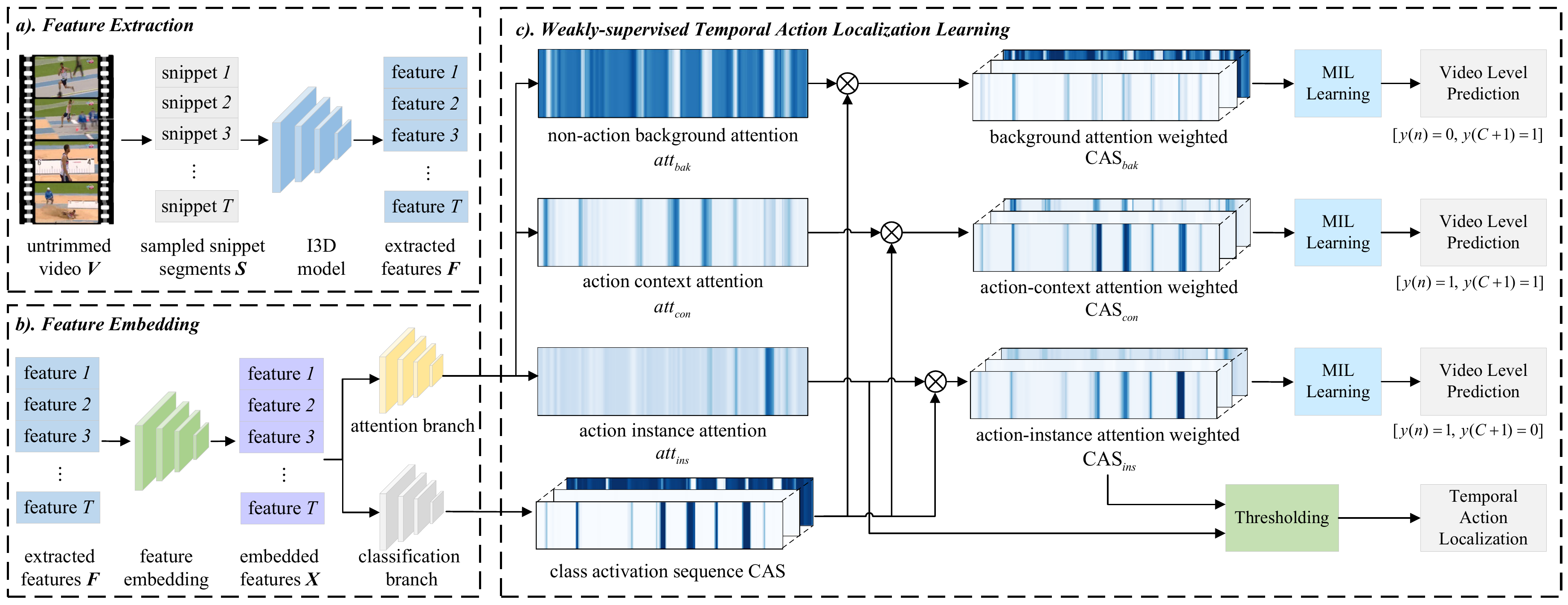}
    \caption{The framework of our proposed ACM-Net, which consists of three parts, i.e. pre-trained feature extraction, video feature embedding, and weakly-supervised temporal action localization guided by action context modeling. We first apply the pre-trained model to extract video snippets level spatial and temporal features, and then employ the feature embedding module to map the pre-trained feature to task-specific feature space. Therefore, to suppress the ambiguous action context frames, we propose a three-branch attention module and multiply the obtained attention values by the raw class activation sequences CAS to obtain the weighted CAS of the corresponding branch, and then employ a multi-instance learning mechanism to learn and model action instance features under supervision with only video-level labels.
    }
    \label{fig:framework}
\end{figure*}

\par In this section, we first define the formulation of weakly-supervised temporal action localization (W-TAL), then present our action-context modeling network (ACM-Net) in detail, and thereafter, introduce the training and inference details. The overview architecture of our ACM-Net is illustrated in Fig.~\ref{fig:framework}.

\subsection{Problem Formulation}
\label{sec:problem_formulation}
\par Assume we are given an untrimmed video $V$, which contains multiple action instances $\{\psi_i = (t^s_i, t^e_i, c_i)\}_{i=1}^{N_{\psi}}$, where $N_{\psi}$ is the number of action instances, $t^s_i$ and $t^e_i$ denotes the start and end time of action instance $\psi_i$, and $c_i \in \mathbb{R}^{C}$ represents the class category. The goal of temporal action localization is to detect all action instances $\{\hat{\psi}_i = (\hat{t}^s_i, \hat{t}^e_i, \hat{c}_i, \hat{\phi}_i)\}_{i=1}^M$, where $\hat{\phi_i}$ denotes the confidence score of action instance $\hat{\psi_i}$.
\par Different from the fully-supervised temporal action localization task, during training, the action instance annotations are available. For the W-TAL task, we can only access the one-hot video-level category label $y = \{0, 1\} \in \mathbb{R}^{C+1}$, where $C$ is the number of action classes and $C+1$ represents non-action background class.

\subsection{Action Context Modeling Network}
\subsubsection{Feature Extraction}
\par Following recent W-TAL methods~\cite{WTALC, AUTOLOC, BASNet, A2CLPT}, for a given untrimmed video $V$, we first divide it into non-overlapping snippets based on a pre-defined sampling ratio, and then apply pre-trained networks to extract snippet-level features. Since different videos vary in temporal length, during training, we utilize interpolation operation to keep all training video have the same time dimension $T$, i.e, for each video we keep the video snippets as $S=\{s(t)\}^T_{t=0}$. As for snippet $s(t)$ feature extraction, we utilize the spatial stream (RGB) and the temporal stream (optical flow) to encode the static scenes feature $F^{rgb}(t) \in \mathbb{R}^D$ and the motion features $F^{flow}(t) \in \mathbb{R}^D$, respectively. Thereafter, we concatenate the static scenes feature $F^{rgb}(t)$ and the motion features $F^{flow}(t)$ together to form the snippet feature $F(t) = [F^{rgb}(t), F^{flow}(t)] \in \mathbb{R}^{2D}$. Afterwards, we stack all snippets feature to form the video pre-trained feature $F \in \mathbb{R}^{T\times2D}$.

\subsubsection{Feature Embedding}
\par Since the extracted features $F$ are not trained from scratch for the W-TAL task, in order to map the extracted video feature $F$ to task-specific feature space, we introduce a feature embedding module. Concretely, we apply a set of convolution layers and non-linear activation functions to map the original video feature $F \in \mathbb{R}^{T\times 2D}$ to task-specific video feature $X\in \mathbb{R}^{T\times 2D}$. Formally, we can denote the feature embedding module as following:
\begin{equation}
    X = \mathrm{ReLU}(\mathrm{Conv}(F, \theta_{embed}))
\end{equation}
where $\theta_{embed}$ denotes the trainable convolution parameters of feature embedding layer and $\mathrm{ReLU}$ is the non-linear activation function we applied in this module.

\subsubsection{Action Class Activation Modeling}
\par To localize action instances in the untrimmed video $V$, based on the embedded video feature $X$, we first apply a snippet-level action classification branch to obtain the Class Activation Sequence (CAS). Even though this CAS can not suppress those ambiguous semantics action contexts well, it is capable of suppressing action-related and non-action-related frames. We set this CAS as an initial indicator for action instances.  Concretely, we apply a multi-layer inception to project the embedded feature to action class category space. The output is $\Phi \in \mathbb{R}^{T\times (C+1)}$, which denotes the classification logit of each action class over time.  Formally, we can express the action class activation branch as follows:
\begin{equation}
    \Phi = \mathrm{MLP}(X, \theta_{cas})
\end{equation}
where $\theta_{cas}$ denotes the trainable operation parameters of the action class activation branch.

\subsubsection{Action Context Attention Modeling}
\label{sec:action_context_modeling}
\par As we presented in Fig.~\ref{fig:motivation}, in addition to highly discriminative action instances frames and non-action background frames, there are many ambiguous frames such as ambiguous action-related background scenes frames or ambiguous incomplete action frames. For simplicity, in this paper, we denote all those ambiguous action-related context frames as action-contexts. 
\par To realize the separation of action instances and contexts from the initial CAS, we first introduce a three-branch snippet-level action attention module to detect class-agnostic action instances, semantically ambiguous contexts, and non-action background frames. Specifically, we apply a single convolution layer and softmax function to measure the likelihood of each snippet containing action instance, action context or non-action background. The output of the three-branch attention module is $A = \{(att_{ins}(t), att_{con}(t), att_{bak}(t))\}^T_{t=0} \in \mathbb{R}^{T\times 3}$, where $att_{ins}(t)$, $att_{con}(t)$, $att_{bak}(t)$ indicates the likelihood of snippet $s(t)$ being action instance, action context or background scenes, respectively. Formally, we denote the three-branch snippet-level action attention module as following:
\begin{equation}
    A = \mathrm{Softmax}(\mathrm{Conv}(X, \theta_{att}))
\end{equation}
where $\theta_{att}$ denotes the trainable convolution layer parameters of the three-branch action-context attention branch.
\par Then based on the obtained attention values, in order to discriminate action instances, contexts and action background frames, we build new three-branch class activation sequences $\mathrm{CAS}_{ins}$, $\mathrm{CAS}_{con}$ and $\mathrm{CAS}_{bak}$, respectively. For simplicity, we present the expression of $\mathrm{CAS}_{ins}$ as:
\begin{equation}
    \mathrm{CAS}_{ins}(t) = att_{ins}(t) * \mathrm{CAS}(t)
\end{equation}
where $\mathrm{CAS}_{ins} \in \mathbb{R}^{T\times(C+1)}$ still presents the class activation scores for each snippet, but it can suppress those ambiguous action context snippets activation scores and still keep high values for action instance snippets. Similarly, for the $CAS_{con}$, it can ignore those action-instance frames and focus on action-related context snippets. And for the $CAS_{bak}$, the weighted class activation sequence will also pay more attention to those non-action background snippets.

\subsubsection{Multiple-Instance Learning}
\par As we presented in section~\ref{sec:problem_formulation}, for the W-TAL task, we can only access the video-level action class label during training. Following recent works~\cite{WTALC, MAAN, BASNet, EMMIL}, we apply the Multiple-Instance Learning (MIL) mechanism~\cite{MIL} to obtain the video-level classification scores. Specifically, in MIL, there are two bags for individual samples, namely positive and negative bags. A positive bag contains at least one positive instance and a negative bag contains no positive instance. The goal of MIL is to distinguish each instance to be positive or negative, besides classifying a bag.
\par In this case, we consider the untrimmed video $V$ as a bag of video snippets, and each snippet instance is represented by the corresponding class activation score. In order to measure the loss of each CAS, we aggregate the top-$k$ action classification scores along with all video snippets for each action category and then average them to build the video-level class activation score. Formally, 
\begin{equation}
    \phi_c(V) = \frac{1}{k} \max_{\begin{subarray}\hat{\Phi}_{n;c} \subset \Phi[:, c]\\ |\hat{\Phi}_{n;c}| = k\end{subarray}} \sum_{\forall \phi \in \hat{\Phi}_{n;c}} \Phi
\end{equation}
where $\hat{\Phi}_{n;c}$ is the subset containing $k$ snippets action classification scores for class $c$  and $k$ is a hyper-parameter proportional to the video snippets length $T$, i.e., $k = \max(1, T//r)$, and $r$ is a pre-defined parameter. 
\par Thereafter, we apply a softmax function to the aggregated averaged top-$k_{act}$ scores to obtain the video-level action probability for each action class:
\begin{equation}
    p_c(V) = \frac{\exp(\phi_c(V))}{\sum_{c^{\prime}=1}^{C+1}{\exp(\phi_{c^{\prime}}(V))}}
\end{equation}
where $p_c(V) \in \mathbb{R}^{C+1}$ denotes the probability of video $V$ contains action class $c$.

\par As we presented in section~\ref{sec:action_context_modeling}, to separate action instances, action context and background snippets, we have build three new class activation sequences $\mathrm{CAS}_{ins}$, $\mathrm{CAS}_{con}$, $\mathrm{CAS}_{bak}$ based on the initial CAS and three-branch attention values. Therefore, for the action-instance attention weighted $\mathrm{CAS}_{ins}$, by applying above MIL mechanism, we can obtain the video-level action probability distribution $p_c^{ins}(V)$. Similarly, for the $\mathrm{CAS}_{con}$ and $\mathrm{CAS}_{bak}$, we can acquire $p_c^{con}(V)$ and $p_c^{bak}(V)$, respectively.

\par Following previous work~\cite{STPN, WTALC, BASNet, EMMIL, ACSNET}, we apply the cross-entropy loss function between the predicted video-level action probability distribution $p_c(V)$ and the ground truth video action probability distribution $y_c(V)$ to classify different action classes in a video. Specifically, for the $\mathrm{CAS}_{ins}$, we can formulate the classification cross-entropy loss as:
\begin{equation}
    \mathcal{L}_{cls}^{ins} = \sum_{c=1}^{C+1} -y_c^{ins}(V) \log(p_c^{ins}(V))
\end{equation}
where $y_c^{ins}$ is the normalized video-level label for the $c$-th class of the video $V$. We set the video-level label $y^{ins} = [y(n)=1, y(C+1)=0]$, since the with the action-instance attention weighting, in $\mathrm{CAS}_{ins}$,  non-action background and ambiguous action context snippets have been suppressed. With similar manner, we can obtain the cross-entrophy loss $\mathcal{L}_{cls}^{con}$ and $\mathcal{L}_{cls}^{bak}$, respectively. Note that, we set the video-level label $y^{con} = [y(n) = 1, y(C+1) = 1]$, $y^{bak} = [y(n) = 0, y(C+1) = 1]$, since with the attention values weighting, $\mathrm{CAS}_{con}$ and $\mathrm{CAS}_{bak}$ pays more attention on action contexts and background scenes instead of action instance snippets. 

\par After we obtained three video-level label based classification loss, $\mathcal{L}_{cls}^{ins}$,  $\mathcal{L}_{con}^{ins}$,  $\mathcal{L}_{bak}^{ins}$, we can compose the overall classification loss $\mathcal{L}_{cls} $ as:
\begin{equation}
    \mathcal{L}_{cls} = \mathcal{L}_{cls}^{ins} + \mathcal{L}_{cls}^{con} + \mathcal{L}_{cls}^{bak}
\end{equation}


\subsection{Optimization Objects}
\label{sec:add_loss}

\par In addition to the regular classification loss $\mathcal{L}_{cls}$, we apply three additional losses to make the network achieve better performance, 1) Attention-Guide Loss $\mathcal{L}_{gui}$, which is used to constrain the action instance attention weighted $\mathrm{CAS}_{ins}$ to follow the action instance attention. 2) Action Feature Separation Loss $\mathcal{L}_{feat}$ for separating the action instances, action contexts, and background snippets features in feature norm space. And 3) Sparse Attention Loss $\mathcal{L}_{spa}$ for constraining the action-instance and action-context branch pays more attention to those action-related frames. The overall loss function is formulated as follows:
\begin{align}
    &\mathcal{L} = \mathcal{L}_{cls} + \mathcal{L}_{add} \\
    &\mathcal{L}_{add} = \lambda_1\mathcal{L}_{gui} + \lambda_2\mathcal{L}_{feat} + \lambda_3\mathcal{L}_{spa}
\end{align}
where $\lambda_1$, $\lambda_2$ and $\lambda_3$ are three hyper-parameters used to balancing the overall loss items.
\subsubsection{Attention Guide Loss} 
\par Although we have introduced the MIL learning mechanism to build video-level classification loss to make the network classify action instances contained in a video, this manner does not optimize the action classification results at snippet-level, which is not conducive to subsequent precise action temporal localization. To make the action classification branch distinguish action instance snippets from those ambiguous action context frames at snippet-level, in addition to the applied video-level cross-entropy classification loss $\mathcal{L}_{cls}$, we introduce the attention guide loss. We set the action instance attention sequence $att_{ins}$ as a binary indicator for each video snippet, and use it to guide the weighted $\mathrm{CAS}_{ins}$ suppress action context and background snippets at snippet-level. Specifically, we compose the attention guide loss $\mathcal{L}_{gui}$ as:
\begin{equation}
    \mathcal{L}_{gui} = \frac{1}{T}\sum_{t=0}^{T} |1 - p^{ins}_{C+1}(t) - att_{ins}(t)|
\end{equation}
where $p^{ins}(t)$ is the predicted snippet-level action probability distribution with softmax function applied on the weighted $\mathrm{CAS}_{ins}$, $p_{C+1}^{ins}(t)$ denotes the likelihood of snippet $s(t)$ not containing action instances, and $att_{ins}(t)$ is the action instance attention branch values at snippet $s(t)$. By minimizing $\mathcal{L}_{gui}$, we can guide the network optimize the class activation sequence at snippet-level.

\subsubsection{Action Feature Separation Loss}
\par To make the embedded video snippets features more distinguishable from action instance, action context, and background features, we introduce the action feature separation loss $\mathcal{L}_{feat}$ at feature norm space. Specifically, the $\mathcal{L}_{feat}$ is defined as:
\begin{align}
    \mathcal{L}_{feat}^{ins} &= \max(0, m - ||X_{cls}|| + ||X_{con}||) \nonumber\\
    \mathcal{L}_{feat}^{con} &= \max(0, m - ||X_{con}|| + ||X_{bak}||) \nonumber\\
    \mathcal{L}_{feat}^{bak} &= ||X_{bak}|| \nonumber\\
    \mathcal{L}_{feat} &= (\mathcal{L}_{feat}^{ins} + \mathcal{L}_{feat}^{con} + \mathcal{L}_{feat}^{bak})^2
\end{align}
where $||\cdot||$ is the feature norm function, $m$ is a pre-defined feature norm separation margin hyper-parameter. And $X_{ins}$, $X_{con}$, $X_{bak}$ are the video-level action instance, action context, and background features, which are built based on the aforementioned top-$k$ mechanism. For simplicity, we present the $X_{cls}$ formulation as follows:
\begin{align}
    X_{cls} &= \frac{1}{k_{ins}} \sum^{\mathrm{idxs} = \mathrm{argsort}(att_{ins})[:k_{ins}]}_{i \in \mathrm{idxs}} X(i)
\end{align}
where $\mathrm{argsort}$ is a function that returns the indices that would sort an array with descending order, $k_{ins}$ is a pre-defined hyper-parameter proportional to the video snippets length $T$ like the aforementioned $k$. In a similar manner, we can obtain the video level action context features $X_{con}$ and background features $X_{bak}$, individually.

\subsubsection{Sparse Attention Loss}
\par Following~\cite{STPN, WTALC, ACSNET}, we also introduce the sparsity attention loss $L_{sparse}$ to constrain the network optimization process, which is based on one assumption that an action can be recognized with a sparse subset of key-snippets in a video. Formally, the $L_{spar}$ is defined as:
\begin{equation}
    \mathcal{L}_{spa} = \frac{1}{T}\sum_{t=1}^{T} att_{ins}(t) + att_{con}(t)
\end{equation}

\subsection{Temporal Action Localization}
\par During the inference, given a test video, we first apply the action instance attention based video-level action probability distribution $p_{c}^{ins}(V)$ to classify the test video based on a pre-defined classification threshold $\bar{p}$. And then we apply threshold strategy on the action instance attention weighted classification sequence $\mathrm{CAS}_{ins}$ and the action instance attention sequence $att_{ins}$ to localize actions. Let $\{(\hat{t}^s_i, \hat{t}^e_i, \hat{c}_i, \hat{\phi}_i)\}$ denotes the detected action instances, like the previous works~\cite{BASNet, A2CLPT, ACSNET}, we apply the Outer-Inner-Contrastive function proposed in~\cite{AUTOLOC} to obtain each detected action instance confidence score $\hat{\phi}_i$. Concretely, the confidence score is defined as:
\begin{align}
    &v = (1 - \alpha) \cdot \mathrm{CAS}_{ins} + \alpha \cdot att_{ins} \nonumber \\
    &\hat{\phi}_i = \frac{\int_{\hat{t}^s_i}^{\hat{t}^e_i} v_{\hat{c}_i}(t)}{\hat{t}^e_i - \hat{t}^s_i} - \frac{\int_{\hat{t}^s_i - \hat{t}^l_i}^{\hat{t}^s_i} v_{\hat{c}_i}(t) + \int_{\hat{t}^e_i}^{\hat{t}^e_i + \hat{t}^l_i} v_{\hat{c}_i}(t)}{2\hat{t}^l_i}
\end{align}
where $\alpha$ is a hyper-parameter coefficient used to combine the $\mathrm{CAS}_{ins}$ and attention values $att_{ins}$, $\hat{t}_i^s$ and $\hat{t}_i^e$ are the
temporal boundary of the detected action instance, $\hat{t}^l_i = (\hat{t}^e_i - \hat{t}^s_i) / 5$ represents the inflated contrast area, and $\hat{c}_i$ denotes the corresponding action instance category. Note that to increase proposals pool, we apply multiple thresholds on the $\mathrm{CAS}_{ins}$ and $att_{ins}$, and then we perform non-maximum-suppression (NMS) to remove overlapped action instances proposals.

\section{Experiments}
\par In this section, we first introduce the datasets and our implementation details about our network, and then compare our methods with the state-of-the-art methods. At last, we apply a set of ablation studies to evaluate the effectiveness of each module component. 

\begin{table*}[ht]
	\centering
	\caption{Temporal action localization performance comparison on the THUMOS-14 dataset. Avg means average mAP from IoU 0.1 to 0.7 with 0.1 increments. Recent works in both fully-supervised and weakly-supervised settings are reported. Our method outperforms the state-of-the-art weakly-supervised methods and even can achieve comparable performance with recent fully-supervised methods when t-IoU$\le$ 0.5. Even though SF-Net~\cite{SFNet} introduces stronger supervision of action instances, we can still achieve better performance. $^*$ denotes the reported performance not containing post-processing.}
	\scalebox{1.0}{
		\begin{tabular}{clcccccccccc}
			\toprule
			\multirow{2}{*}{Supervision}&\multirow{2}{*}{Method}&\multicolumn{10}{c}{mAP@t-IoU(\%)}\\
			\cmidrule{3-12} & &0.10 &0.20 &0.30 &0.40 &0.50 &0.60 &0.70 &Avg[0.1-0.5] &Avg[0.3-0.7] &Avg\\
			\midrule
			\multirow{6}{*}{Full action label}
			&SSN (ICCV 2017)~\cite{SSN}    &66.0 &59.4 &51.9 &41.0 &29.8 &- &- &49.6 &- &-  \\
            &BSN (ECCV 2018)~\cite{BSN}    &- &- &53.5 &45.0 &36.9 &28.4 &20.0 &- &36.8 &-\\
            &BMN (ICCV 2019)~\cite{BMN}    &- &- &56.0 &47.4 &38.8 &29.7 &20.5 &- &38.5 &-  \\
            &G-TAD$^*$ (CVPR 2020)~\cite{GTAD}    &\bf{66.1} &\bf{64.2} &54.5 &47.6 &40.2 &30.8 &\bf{23.4} &\bf{54.5} &39.3 &\bf{46.7}  \\
            &BSN++ (AAAI 2021)~\cite{BSN++}   &- &- &\bf{59.9} &\bf{49.5} &\bf{41.3} &\bf{31.9} &22.8 &- &\bf{41.1} &-  \\
			\midrule
			Weak action label &SF-Net (ECCV 2020)~\cite{SFNet}   &\bf{71.0} &\bf{63.4} &\bf{53.2} &\bf{40.7} &\bf{29.3} &\bf{18.4} &\bf{9.6} &\bf{51.5} &\bf{30.2} &\bf{40.8}  \\
            \midrule 
			\multirow{14}{*}{Weak video label}
			&Hide-and-Seek (ICCV 2017)~\cite{HIDE}          &36.4 &27.8 &19.5 &12.7 &6.8 &- &- &- &20.6 &-   \\
            &UntrimmedNet (CVPR 2017)~\cite{UntrimmedNet}   &44.4 &37.7 &28.2 &21.2 &13.7 &- &- &- &29.0 &-  \\
            &STPN (CVPR 2018)~\cite{STPN}    &52.0 &44.7 &35.5 &25.8 &16.9 &9.9 &4.3 &35.0 &18.5 &26.4  \\
            &AutoLoc (ECCV 2018)~\cite{AUTOLOC}    &- &- &35.8 &29.0 &21.2 &13.4 &5.8 &- &21.0 &-  \\
            &W-TALC (ECCV 2018)~\cite{WTALC}    &55.2 &49.6 &40.1 &31.1 &22.8 &- &7.6 &39.8 &- &-  \\
            &IWO-Net (TIP 2019)~\cite{IWONet} &57.6	&48.9 &38.9	&29.3 &20.5 &- &- &39.0 &- &-\\
            &MAAN (ICLR 2019)~\cite{MAAN}   &59.8 &50.8 &41.1 &30.6 &20.3 &12.0 &6.9 &40.5 &22.2 &31.6  \\
            &TSM (ICCV 2019)~\cite{TSM}   &- &- &39.5 &- &24.5 &- &7.1 &- &- &- \\
            &BasNet (AAAI 2020)~\cite{BASNet}   &58.2 &52.3 &44.6 &36.0 &27.0 &18.6 &10.4 &43.6 &27.3 &35.3   \\
            &DGAM (CVPR 2020)~\cite{DGAM}   &60.0 &54.2 &46.8 &38.2 &28.8 &19.8 &11.4 &45.6 &29.0 &37.0  \\
            &EM-MIL (ECCV 2020)~\cite{EMMIL}   &59.1 &52.7 &45.5 &36.8 &30.5 &\bf{22.7} &\bf{16.4} &45.0 &30.4 &37.7  \\
            &A2CL-PT (ECCV 2020)~\cite{A2CLPT}   &61.2 &56.1 &48.1 &39.0 &30.1 &19.2 &10.6 &46.9 &29.4 &37.8  \\
            &ACSNet (AAAI 2021)~\cite{ACSNET}   &- &- &51.4 &42.7 &32.4 &22.0 &11.7 &- &32.0 &-  \\
            &CoLA (CVPR 2021)~\cite{CoLA} &66.2 &59.5 &51.5 &41.9 &32.2 &22.0 &13.1 &50.3 &32.1 &40.9\\  
            &\bf{ACM-Net(Ours)}   &\bf{68.9} &\bf{62.7} &\bf{55.0} &\bf{44.6} &\bf{34.6} &21.8 &10.8 &\bf{53.2} &\bf{33.4} &\bf{42.6}  \\
			\bottomrule
	\end{tabular}}
	\label{table:comparison_thu14}
\end{table*}

\begin{table}[ht]
	\centering
	\caption{Temporal action localization performance comparison on the ActivityNet-1.3 dataset. Avg means average mAP from IoU 0.50 to 0.95 with 0.05 increments. Our method outperforms the state-of-the-art  weakly-supervised methods, and especially at t-IoU$=0.5$, we can obtain more than $3\%$ improvement.}
	\scalebox{0.88}{
		\begin{tabular}{clcccc}
			\toprule
			\multirow{2}{*}{Supervision}&\multirow{2}{*}{Method}&\multicolumn{4}{c}{mAP@t-IoU(\%)}\\
			\cmidrule{3-6} & &0.50 &0.75 &0.95 &Avg\\
			\midrule
			\multirow{6}{*}{Full}
			&SSN (ICCV 2017)~\cite{SSN}   &39.1 &23.5 &5.5 &24.0\\
            &BSN (ECCV 2018)~\cite{BSN}   &46.5 &30.0 &8.0 &30.0\\
            &BMN (ICCV 2019)~\cite{BMN}   &50.1 &34.8 &8.3 &33.9\\
            &P-GCN (CVPR 2019)~\cite{PGCN}   &48.3 &33.2 &3.3 &31.1\\
            &G-TAD (CVPR 2020)~\cite{GTAD}   &50.4 &34.6 &\bf{9.0} &34.1\\
            &BSN++ (AAAI 2021)~\cite{BSN++}  &\bf{51.3} &\bf{35.7} &8.3 &34.9\\
			\midrule
			\multirow{7}{*}{Weak}
            &STPN (CVPR 2018)~\cite{STPN}    &29.3 &16.9 &2.6 &-\\
            &MAAN (ICLR 2019)~\cite{MAAN}   &33.7 &21.9 &5.5  &-\\
            &TSM (ICCV 2019)~\cite{TSM}   &30.3 &19.0 &4.5  &-\\
            &BasNet (AAAI 2020)~\cite{BASNet}  &34.5 &22.5 &4.9 &22.2\\
            &A2CL-PT (ECCV 2020)~\cite{A2CLPT} &36.8 &22.5 &5.2 &22.5\\
            &ACSNet (AAAI 2021)~\cite{ACSNET} &36.3 &\bf{24.2} &5.8 &23.9\\
            &\bf{ACM-Net(Ours)}  &\bf{40.1} &\bf{24.2} &\bf{6.2} &\bf{24.6}\\
			\bottomrule
	\end{tabular}}
	\label{table:comparison_act13}
\end{table}

\subsection{Experiments Setting}

\subsubsection{Datasets}
\par We perform extensive experiments on two large-scale temporal action localization datasets THUMOS-14~\cite{thu14_dataset} and ActivityNet1.3~\cite{Activitynet}.
\par THUMOS-14~\cite{thu14_dataset}, which contains 200 untrimmed validation videos and 213 untrimmed test videos with precise temporal action boundary annotations belonging to 20 action class categories. On average, each video contains 15.4 action instances and more than 70\% frames are ambiguous action contexts or non-action background scenes. In addition, the video length varies from a few seconds to more than one hour, which makes it very challenging especially for weakly-supervised temporal action localization. Following previous work, we apply the validation videos for training and the test videos for testing.
\par ActivityNet1.3~\cite{Activitynet}, which contains 10024 untrimmed training videos, 4926 untrimmed validation untrimmed videos, and 5044 videos for testing whose action instance labels are withheld. The action instance class categories involved in this dataset are 200. On average, each video contains 1.6 action instances and about 36\% frames are ambiguous action contexts or non-action background scenes. For a fair comparison, same as the previous work, we also utilize training videos for training and report experiments results on the validation videos.

\subsubsection{Evaluation Metrics}
\par We use the mean Average Precision (mAP) with different temporal Intersection over Union (t-IoU) thresholds to evaluate our weakly-supervised temporal action localization performance, which denotes as mAP@t-IoU. Specifically, the t-IoU thresholds for THUMOS-14 is [0.1:0.1:0.7] and for ActivityNet is [0.5:0.05:0.95].

\subsubsection{Implementation Details}

\par For the feature extraction, we first sample RGB frames at 25 fps for each video and apply the TV-L1 algorithm~\cite{tvl_optical_flow} to generate optical flow frames. Then, we divide each video into non-overlapping snippets with consecutive 16 frames. Thereafter, we perform the I3D networks~\cite{I3D} pre-trained on the Kinetics dataset~\cite{kinetics_dataset} to obtain the video feature $F$. Note that, for a fair comparison, we do not introduce any other feature fine-tuning operations to the pre-trained I3D model.

\par For the training process on the THUMOS-14 dataset, we set the training video batch size to 16, and apply the Adam optimizer~\cite{adam} with learning rate $10^{-4}$ and weight decay $5\times10^{-4}$. We set the video snippets length $T=750$, and the top-$k$ number for action instance branch $k_{ins} = T // r_{ins}$, for action context branch $k_{con} = T // r_{con}$, for action background branch $k_{bak} = T // r_{bak}$. According to parameter fine-tuning, we set $r_{ins} = 8$, $r_{con} = r_{bak} = 3$, $\lambda_1 = 2\times10^{-3}$, $\lambda_2 = 5\times10^{-5}$, $\lambda_3 = 2\times10^{-4}$,  $\alpha = 0$. For the action instance proposals generation, we set the thresholds from 0.15 to 0.25 with step 0.05, to remove overlap proposals, we perform NMS with an t-IoU threshold of 0.50.

\par For the training process on the ActivityNet-1.3 dataset, we set the training video batch size to 64, the optimizer~\cite{adam} learning rate to $10^{-4}$ and weight decay $0.001$. Considering that most video length varies from a few seconds to several minutes, much shorter than the THUMOS-14 dataset, we set the video snippets length $T=75$. According to parameter fine-tuning, we set $r_{ins} = 2$, $r_{con} = r_{bak} = 10$, $\lambda_1 = 5\times10^{-3}$, $\lambda_2 = 1\times10^{-5}$, $\lambda_3 = 0$,  $\alpha = 0.5$. For the action instance proposals generation, we set the thresholds from 0.01 to 0.02 with step 0.005, to remove overlap proposals, we apply NMS with an t-IoU threshold of 0.90. 
\par All the experiments are evaluated with PyTorch-1.7~\cite{pytorch} on the RTX-3090 platform.

\subsection{Comparison with the State-of-the-arts Methods}
\par We compare our proposed network with current existing fully-supervised and weakly-supervised temporal action localization methods. 

\par Table.~\ref{table:comparison_thu14} compares our method with current fully and weakly-supervised based temporal action localization methods on THUMOS-14 dataset. As shown, by introducing the action context modeling mechanism, our method can achieve new state-of-the-art performances under weak video label constraints.
Even compared with SF-Net~\cite{SFNet} that introduces more strong supervision (for each action instance, the SF-Net introduces a temporal point annotations), we can still achieve better performance. Furthermore, it can be observed that our methods even can achieve comparable performance with recently fully-supervised methods when t-IoU $\le 0.5$, even though we do not have access to more detailed and specific action instances annotations during training.

\par Table.~\ref{table:comparison_act13} presents the performance comparison on the ActivityNet-1.3 dataset. As shown, our method can also achieve new state-of-the-art performance under the weakly-supervised assumption. However, the performance improvement is not as significant as the THUMOS-14 dataset.
This may be because the ground truth results of the action instances in ActivityNet-1.3 are not as precise as the THUMOS-14 dataset, we found that the action instances annotations on ActivityNet-1.3 are more prone to contain some ambiguous action context frames.

\subsection{Ablation Study and Analysis}
\begin{table}[t]
	\centering
	\caption{Ablation study of the effectiveness of our proposed action context modeling mechanism on the THUMOS-14 dataset. Avg means average mAP from t-IoU 0.1 to 0.7 with 0.1 increments.}
	\scalebox{0.85}{
		\begin{tabular}{c|ccc|c|ccccc}
			\toprule
			\multirow{2}{*}{Exp} &\multicolumn{3}{c|}{$\mathcal{L}_{cls}$}& \multirow{2}{*}{$\mathcal{L}_{add}$}  &\multicolumn{5}{c}{mAP@t-IoU(\%)}\\
			\cmidrule{2-4} \cmidrule{6-10} & $\mathcal{L}_{cls}^{ins}$ &$\mathcal{L}_{cls}^{con}$ &$\mathcal{L}_{cls}^{bak}$ & &0.10 &0.30 &0.50 &0.70 &Avg\\
			\midrule
			1&\cmark &\xmark &\xmark &\xmark  &49.9 &32.9 &16.6 &5.3 &26.1 \\
            2&\cmark &\xmark &\cmark &\xmark  &55.9 &41.9 &23.0 &7.1 &32.0 \\
            3&\cmark &\cmark &\xmark &\xmark  &\bf{67.4} &\bf{50.8} &\bf{31.5} &\bf{10.8} &\bf{40.3}\\
            4&\cmark &\cmark &\cmark &\xmark  &65.6 &49.4 &29.6 &10.0 &38.8 \\
            \midrule
            5&\cmark &\cmark &\cmark &\cmark  &\bf{68.9} &\bf{55.0} &\bf{34.6} &\bf{10.8} &\bf{42.6} \\
			\bottomrule
	\end{tabular}}
	\label{table:action_context_ablation_thu14}
\end{table}

\begin{table}[t]
	\centering
	\caption{Ablation study of the effectiveness of our proposed action context modeling mechanism on the ActivityNet-1.3 dataset. Avg means average mAP from t-IoU 0.50 to 0.95 with 0.05 increments.}
	\scalebox{0.90}{
		\begin{tabular}{c|ccc|c|cccc}
			\toprule
			\multirow{2}{*}{Exp} &\multicolumn{3}{c|}{$\mathcal{L}_{cls}$}& \multirow{2}{*}{$\mathcal{L}_{add}$}  &\multicolumn{4}{c}{mAP@t-IoU(\%)}\\
			\cmidrule{2-4} \cmidrule{6-9} & $\mathcal{L}_{cls}^{ins}$ &$\mathcal{L}_{cls}^{con}$ &$\mathcal{L}_{cls}^{bak}$ & &0.50 &0.75 &0.95 &Avg\\
			\midrule
			1&\cmark &\xmark &\xmark &\xmark  &34.5  &21.0  &4.9  &21.5 \\
            2&\cmark &\cmark &\xmark &\xmark  &\bf{39.1}  &\bf{23.4}  &\bf{6.0} &\bf{23.9}\\
            3&\cmark &\cmark &\cmark &\xmark  &39.1  &23.1  &5.8  &23.7 \\
            \midrule
            4&\cmark &\cmark &\cmark &\cmark  &\bf{40.1} &\bf{24.2} &\bf{6.2} &\bf{24.6} \\
			\bottomrule
	\end{tabular}}
	\label{table:action_context_ablation_act13}
\end{table}

\subsubsection{Effectiveness of action context modeling}
\par To demonstrate the effectiveness of our proposed action context branch modeling mechanism for weakly-supervised temporal action localization, we apply extensive ablation experiments on THUMOS-14 dataset. The results are summarized in Table.~\ref{table:action_context_ablation_thu14}. For simplicity, we denote the additional loss constraints  introduced in Section~\ref{sec:add_loss} as $L_{add}$.
\par As shown in the table, compared with baseline methods not introducing action context modeling mechanism (Exp $1, 2$), our proposed action context modeling mechanism is greatly beneficial for temporal action localization (Exp $3, 4$). Specifically, without bells and whistles, we can achieve more than $14\%$ and $6\%$ average performance gain. We attribute this huge performance gain to the fact that it is unreasonable to force the network to group all the action instances, contexts, and non-action background snippets into one category since they do not share any common semantics. It is noteworthy that the experiment $3$ only based $\mathcal{L}_{cls}^{ins}$ and $\mathcal{L}_{cls}^{con}$ constrains can achieve better performance compared to the experiment $4$ introduced the $\mathcal{L}_{cls}^{bak}$. This may be attributable to the fact that the network cannot make a clearer distinction between action instances, contexts, and background fragments without introducing additional constraints $L_{add}$, whose effectiveness will be evaluated in the next subsection.

\par Besides, we can observe that on the THUMOS-14 dataset only based on the $\mathcal{L}_{cls}^{ins}$ and $\mathcal{L}_{cls}^{con}$ constrains, without introducing any other tricks or losses, we can achieve a performance comparable to current SOTA methods, which further proves the effectiveness of our action context modeling mechanism.
\par We can also draw the same conclusion on the ActivityNet-13 dataset that we can achieve the current SOTA performance only by introducing the action context modeling mechanism, the results of the ablation study on ActivityNet-1.3 are presented in Table.~\ref{table:action_context_ablation_act13}.

\subsubsection{Effectiveness of additional loss}
\begin{table}[t]
	\centering
	\caption{Evaluation the effectiveness of additional loss functions on the THUMOS-14 dataset. }
	\scalebox{0.85}{
		\begin{tabular}{c|ccc|c|ccccc}
			\toprule
			\multirow{2}{*}{Exp} &\multicolumn{3}{c|}{$\mathcal{L}_{add}$}& \multirow{2}{*}{$\mathcal{L}_{cls}$}  &\multicolumn{5}{c}{mAP@t-IoU(\%)}\\
			\cmidrule{2-4} \cmidrule{6-10} &$\mathcal{L}_{gui}$ &$\mathcal{L}_{feat}$ &$\mathcal{L}_{spa}$ & &0.10 &0.30 &0.50 &0.70 &Avg\\
			\midrule
			1&\xmark &\xmark &\xmark &\cmark  &65.6 &49.4 &29.6 &10.0 &38.8 \\
			2&\cmark &\xmark &\xmark &\cmark   &66.4 &51.5 &32.2 &11.0 &40.5 \\
            3&\xmark &\cmark &\xmark &\cmark   &66.4 &51.2 &31.1 &10.3 &39.9 \\
            4&\xmark &\xmark &\cmark &\cmark   &64.8 &48.9 &30.3 &10.7 &38.7 \\
            5&\cmark &\cmark &\xmark &\cmark   &67.4 &53.1 &33.8 &\bf{11.2} &41.8 \\
            6&\cmark &\cmark &\cmark &\cmark  &\bf{68.9} &\bf{55.0} &\bf{34.6} &10.8 &\bf{42.6} \\
			\bottomrule
	\end{tabular}}
	\label{table:addition_loss_ablation_thu14}
\end{table}

\par Following previous works, we have introduced additional constraints $\mathcal{L}_{add}$ to further improve the temporal action detection performance. To analysis the impact of each additional loss constraint on the final temporal action detection performance, we make extensive ablation experiments on THUMOS-14 dataset. The results are collated in the Table.~\ref{table:addition_loss_ablation_thu14}.

\par As the table presents, both the $\mathcal{L}_{gui}$ and $\mathcal{L}_{feat}$ are beneficial for boosting the temporal action localization performance. For the $\mathcal{L}_{gui}$, we can obtain $1.7\%$ average mAP performance gain, since this constrain can facilitate the network to minimize the difference between $\mathrm{CAS}_{ins}$ and $att_{ins}$ at snippet-level.
As regards the $\mathcal{L}_{feat}$, since it is beneficial to separate action instances, action contexts, and non-action background snippets, we can arrive $1.1\%$ average mAP performance boost compared to the baseline experiment. However, we found that the $\mathcal{L}_{spa}$ can not stand alone for improving the performance, possibly because directly minimizing the temporal attention values may cause the network to pay more attention to those action salient snippets and ignore those less discriminative action snippets. When we combine the $\mathcal{L}_{gui}$, $\mathcal{L}_{feat}$ and $\mathcal{L}_{spa}$ together as $\mathcal{L}_{add}$, we can derive $3.6\%$ performance improvement and boost our method to a  new state-of-the-art. 

\begin{figure*}[!h]
	\centering
    \includegraphics[width=0.98\textwidth]{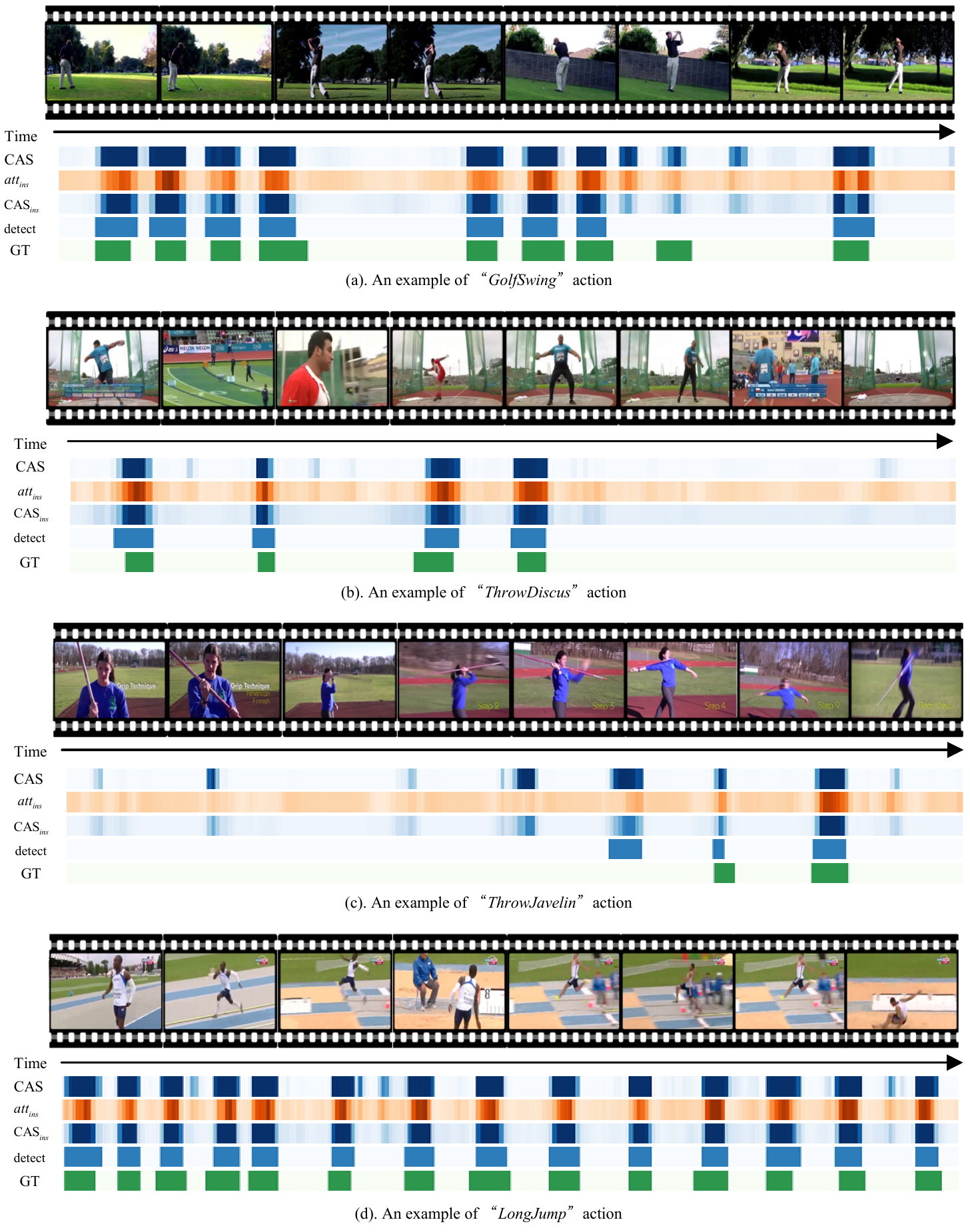}
	\caption{Qualitative result visualization on the THUMOS-14 dataset. From the above qualitative results, we can conclude that our proposed action context modeling mechanism is greatly beneficial to suppress ambiguous action context frames and help us achieve more precise temporal action localization results. However, we can note that this mechanism is not perfect and that it may sometimes suppress the true action instance frames. The possible reason for this is that those action instance frames may not be distinctive, but rather ambiguous, similar to action context frames.}
	\label{fig:qualitative_result}
\end{figure*}

\subsubsection{Analysis on video snippets number $T$}

\par As we mentioned before, natural videos always vary in temporal length, however, under the weakly supervised formulation of the temporal action localization problem, we can only access video-level labels during the training process. Therefore, to achieve parallel optimization of the proposed network, we employ a linear interpolation-based sampling strategy such that all training videos have the same temporal dimension $T$. To analysis the influence of the video sampling snippets number $T$ on the final performance, we carried out a series of experiments, the results are presented in the Table.~\ref{table:sample_T_ablation_thu14}. 

\par From the results in the table, we can conclude that $T$ is not linearly related to the growth in detection performance. When $T$ is small, increasing $T$ can lead to significant performance gains (e.g. we can obtain $5.1\%$ average mAP boost from $T = 250$ to $T = 500$), but when $T$ exceeds a certain value ($T = 750$ for the THUMOS-14 dataset), detection performance starts to drop again. The probable reason for this is that when $T$ is small, for most videos we cannot achieve full sampling, while when $T$ exceeds a certain value, it leads to oversampling, resulting in the introduction of more ambiguous context snippets during training.

\begin{table}[t]
	\centering
	\caption{Evaluation the influence of video snippets sample number $T$ on the THUMOS-14 dataset.}
	\label{table:sample_T_ablation_thu14}
	\scalebox{0.90}{
		\begin{tabular}{c|c|cccccccc}
			\toprule
			\multirow{2}{*}{Exp} &\multirow{2}{*}{T} &\multicolumn{8}{c}{mAP@t-IoU(\%)}\\
			\cmidrule{3-10} & & 0.10 &0.20 &0.30 &0.40 &0.50 &0.60 &0.70 &Avg\\
			\midrule
			1 &250 &64.4 &56.3 &47.0 &36.6 &26.4 &15.8 &7.4 &36.3\\
			2 &500 &67.2 &61.2 &53.5 &43.4 &32.5 &20.8 &10.9 &41.4\\
            3 &750 &\bf{68.9} &\bf{62.7} &\bf{55.0} &\bf{44.6} &\bf{34.6} &21.8 &10.8 &\bf{42.6}\\
            4 &900 &67.0 &61.9 &54.0 &43.8 &33.6 &21.3 &10.8 &41.8 \\
            5 &1000 &67.7 &61.7 &53.3 &43.5 &32.4 &20.7 &\bf{11.0} &41.5\\
			\bottomrule
	\end{tabular}}
\end{table}

\subsection{Qualitative Results}

\par To further demonstrate and exemplify the effectiveness of our action context modeling mechanism, we present some qualitative results in Fig.~\ref{fig:qualitative_result}. In this figure, the $\mathrm{CAS}$ denotes the class activation sequence originally obtained from the classification branch. $att_{ins}$ denotes the action instance attention values which are applied to suppress those ambiguous action context frames. While $\mathrm{CAS}_{ins}$ represents the class activation sequence calculated from the action instance attention values $att_{ins}$ weighted $\mathrm{CAS}$. 
\par From those qualitative visualization results, we can conclude that our proposed action context modeling mechanism is of great benefit in suppressing ambiguous action context frames, helping to filter those false-positive errors, and obtaining more accurate temporal action localization results. However, this mechanism is not perfect under the problem definition of weak supervision, since we do not have the access to accurate action annotation information during the training process, therefore the mechanism may sometimes suppress and filter out some of the not so significant and discriminative action frames.


\section{Conclusion}
\par In this paper, we propose an action context modeling network termed ACM-Net to achieve the separation among action instances, semantically ambiguous action context, and non-action background frames. The proposed ACM-Net integrates a three-branch attention module which is used to measure the likelihood of each temporal point containing instance, context, or background frames. Based on the three-branch attention values, three-branch class activation sequences (CAS) are then introduced to represent the action instances, action contexts, or background activation scores at each temporal point. We conduct extensive experiments on two popular benchmark datasets the THUMOS-14 and the ActivityNet-1.3 to demonstrate the effectiveness of our ACM-Net. The results show that our ACM-Net can outperform current weakly-supervised state-of-the-art methods and beat those with stronger supervision, and can even achieve performance comparable to those with full supervision. For the future work, we believe the context modeling will be a promising direction for various weakly supervised learning tasks, and explore such mechanism in other related tasks.


%

\section*{Acknowledgment}
\par This work is funded by National Natural Science Foundation of China (No. 61906138), the European Union’s Horizon 2020 Framework Programme for Research and Innovation under the Specific Grant Agreement No. 945539 (Human Brain Project SGA3), and the Shanghai AI Innovation Development Program 2018.

\ifCLASSOPTIONcaptionsoff
  \newpage
\fi



%
\bibliographystyle{IEEEtran}
\bibliography{bib/acmnet}

%








\end{document}